\documentclass[conference, twocolumn]{IEEEtran}
\usepackage{booktabs}

\usepackage{graphicx}
\usepackage{array}

\usepackage[utf8]{inputenc}
\usepackage[T1]{fontenc}

\usepackage{listings}
\usepackage{subcaption}
\usepackage[table,xcdraw]{xcolor}
\usepackage{amsmath}
\usepackage{bbm}
\usepackage{pifont}
\usepackage{amssymb}
\definecolor{mydarkblue}{rgb}{0,0.08,0.65}
\usepackage[colorlinks=true,
    linkcolor=mydarkblue,
    citecolor=mydarkblue,
    filecolor=mydarkblue,
    urlcolor=mydarkblue]{hyperref}
\usepackage{cleveref}
\usepackage{soul}
\usepackage[shortlabels]{enumitem}
\usepackage{url}
\usepackage{color}
\usepackage{tikz}
\usetikzlibrary{positioning}
\usepackage{balance}
\usepackage{multirow}
\usepackage{makecell}
\usepackage{comment}
\usepackage{wrapfig,booktabs}

\usepackage[symbol]{footmisc}
\usepackage{tablefootnote}
\usepackage{tabularx}
\usepackage{placeins}
\usepackage{algorithm}
\usepackage{algpseudocode}
\usepackage{ragged2e}
\usepackage{cuted}
\usepackage{float}
\usepackage{caption}

\usepackage{booktabs}
\usepackage{xcolor}

\usepackage{multicol}

\usepackage{dblfloatfix}
\usepackage{natbib}

\definecolor{codegreen}{rgb}{0,0.6,0}
\definecolor{codegray}{rgb}{0.5,0.5,0.5}
\definecolor{codepurple}{rgb}{0.58,0,0.82}
\definecolor{backcolour}{rgb}{0.95,0.95,0.92}

\usepackage{xcolor}
\usepackage{pgfplots}
\pgfplotsset{compat=1.18}
\usepgfplotslibrary{groupplots}

\pgfplotscreateplotcyclelist{nagcolors}{
{blue!100!white, thick},
{blue!80!white, thick},
{blue!60!white, thick},
{blue!40!white, thick},
{blue!20!white, thick},
}

\pgfplotscreateplotcyclelist{baselinecolors}{
{red!100!white, thick},
{red!80!white, thick},
{red!60!white, thick},
{red!40!white, thick},
{red!20!white, thick},
}

\definecolor{nagblue}{HTML}{0072B2}
\definecolor{nagorange}{HTML}{E69F00}
\definecolor{naggreen}{HTML}{009E73}
\definecolor{nagred}{HTML}{D55E00}
\definecolor{nagpurple}{HTML}{CC79A7}

\pgfplotscreateplotcyclelist{multicolor}{%
    {color=nagblue, thick, solid, mark=none}\\%
    {color=nagorange, thick, solid, mark=none}\\%
    {color=naggreen, thick, solid, mark=none}\\%
    {color=nagred, thick, solid, mark=none}\\%
    {color=nagpurple, thick, solid, mark=none}\\%
}

\makeatletter
\def\blfootnote{\xdef\@thefnmark{}\@footnotetext}
\makeatother

\lstdefinestyle{mystyle}{
  backgroundcolor=\color{backcolour},   commentstyle=\color{codegreen},
  keywordstyle=\color{magenta},
  numberstyle=\tiny\color{codegray},
  stringstyle=\color{codepurple},
  basicstyle=\ttfamily\footnotesize,
  breakatwhitespace=false,
  breaklines=true,
  captionpos=b,
  keepspaces=true,
  numbers=left,
  numbersep=5pt,
  showspaces=false,
  showstringspaces=false,
  showtabs=false,
  tabsize=2,
}

\AtBeginDocument{%
  \providecommand\BibTeX{{%
    \normalfont B\kern-0.5em{\scshape i\kern-0.25em b}\kern-0.8em\TeX}}}

\IEEEoverridecommandlockouts
\begin{document}

\title{Can Scale Save Us From Plasticity Loss in Large Language Models?}
\newcommand{\corr}{\textsuperscript{*}}

\author{J. Fernando Hernandez-Garcia\corr, Tomas Figliolia\corr, Beren Millidge
\\[0.5em]
\textbf{Zyphra}
\\
San Francisco, CA \\
\IEEEauthorblockA{\textsuperscript{*}Corresponding authors: \texttt{fernando@zyphra.com}, \texttt{tom@zyphra.com}}
}
\maketitle
\setcounter{page}{1}

\begin{abstract}\normalfont\mdseries
The loss of plasticity -- the ability of a network to learn new information after having already learned older information -- is a fundamental challenge in creating artificial neural networks capable of continual learning. Although this phenomenon has been known for decades, it has mostly been studied in older, relatively small architectures and rarely in natural-language domains. To determine whether loss of plasticity remains a problem in the modern transformer-based LLM paradigm, we study plasticity loss in GPT-style Transformer models trained on a multilingual continual learning problem. Consistent with prior work, we find evidence of plasticity loss across models ranging from 5M to 314M non-embedding parameters, as measured by deterioration on a held-out Vietnamese probing task. We further find that the onset of plasticity loss follows a predictable scaling law, growing sublinearly with model size.
These results suggest that larger models may delay the measurable effects of plasticity loss, but that increasing parameter count alone is likely to be insufficient to completely prevent it. We also find evidence of plasticity loss under stationary multilingual training, challenging the view that the phenomenon is exclusive to continual learning with abrupt task changes.
Overall, our results suggest that even large Transformer language models trained on natural-language will eventually lose the ability to efficiently adapt to new data after sufficiently long training, in both continual and stationary settings.
\end{abstract}

\section{Introduction}
\label{sec:introduction}

Neural networks trained for extended periods on changing data can gradually lose their ability to learn from new observations---a phenomenon known as \textit{loss of plasticity}~\citep{ash2020on, lyle2022capacity, nikishin2022the, dohare2024loss}. In this work, we operationalize plasticity loss as a degradation in a model's ability to improve on a target distribution under a fixed training budget. This phenomenon represents a fundamental challenge for developing systems capable of continual learning, since such systems must be capable of incorporating new information without losing the ability to adapt. Importantly, loss of plasticity differs from catastrophic forgetting in that plasticity loss is about the network becoming incapable of learning new information, rather than forgetting old information. 

Continual learning is a fundamental frontier for language models, whose usefulness depends on their ability to adapt to new facts, domains, languages, codebases, and user needs. For example, continual learning could help mitigate knowledge cutoffs and allow coding agents to adapt to new repositories without relying entirely on in-context learning via long prompts. Because of this promise, understanding and preventing plasticity loss has become an increasingly active area of research~\citep{kumar2024maintaining, lyle2024disentangling, chung2024parseval, elsayed2024weight, elsayed2024addressing, lewandowski2024curvature, lyle2024normalization, liu2026neural, lillo2026activation}.

While research on plasticity loss has expanded to include a wider range of domains and architectures, there are still few studies of the phenomenon in natural-language settings with large Transformer models, especially at scale. Previous work involving Transformer architectures has either focused on vision domains~\citep{lyle2023understanding, lee2024slow, hernandez2025reinitializing, lewandowski2025learning} or used synthetic text data~\citep{farias2025selfnormalized}, often with networks containing fewer than 20 million parameters. In natural-language settings with large Transformer models, existing evaluations have considered relatively few tasks, making it difficult to characterize the onset and progression of plasticity loss~\citep{cho2026forget}. Thus, it remains unclear whether GPT-style models trained on realistic text streams exhibit plasticity loss, how this effect changes with model scale, and whether it is a practical problem in real-world applications. 

In this paper, we investigate plasticity loss in GPT-style decoder-only Transformer models trained on natural-language data. Unlike the original Transformer architecture of~\citet{vaswani2017attention}, which used post-layer normalization, our models use the pre-layer-normalization architecture common in modern language models~\citep{radford2019language,brown2020language,deepseekai2025deepseekv3technicalreport}. To measure loss of plasticity, we introduce a multilingual next-token prediction problem in which languages are presented one at a time for a fixed number of tokens, and the process cycles through a sequence of languages so that training can be extended indefinitely. We then periodically evaluate how efficiently each model adapts to a held-out Vietnamese probing task that is not included in the continual pretraining sequence.

Our experiments show evidence of plasticity loss in models ranging from 5M to 314M non-embedding parameters, with smaller models exhibiting measurable deterioration earlier. We find that the onset of plasticity loss follows a predictable sublinear, power-law scaling with model size, suggesting diminishing returns from increasing parameter count alone as a mitigation strategy. We also find evidence of plasticity loss under stationary multilingual training, indicating that abrupt task changes, which are common in the plasticity loss literature but uncommon in practice, are not necessary for the phenomenon to emerge. Finally, we measure several correlates of plasticity loss, including parameter growth, dormant units, and collapsed or lazy attention heads, which provide clues for designing methods that maintain plasticity over long training horizons. However, we do not yet manage to find a 'smoking gun' to diagnose and counteract plasticity loss in LLMs. 


\section{Background}
\label{sec:background}

The finding that artificial neural networks can lose their ability to learn after training dates back more than two decades~\citep{smith2000early, ellis2000age}. Recent work has shown that this phenomenon is not merely a pathology of early neural networks, but also arises in modern deep learning systems trained for extended periods~\citep{lyle2022capacity, nikishin2022the, dohare2023maintaining, abbas2023loss}. These observations have motivated a growing body of work aimed at characterizing when plasticity loss occurs, how broadly it appears across architectures and domains, and how it might be prevented.

Loss of plasticity has been observed in feedforward networks~\citep{dohare2024loss, lyle2023understanding, kumar2024maintaining, elsayed2024addressing}, convolutional networks~\citep{abbas2023loss, nikishin2022the, lee2023plastic}, residual networks~\citep{dohare2024loss, lyle2023understanding, lee2024slow}, and Transformer architectures~\citep{lyle2023understanding, lee2024slow, hernandez2025reinitializing, farias2025selfnormalized, cho2026forget, liu2026neural}. In addition, the phenomenon has been observed across multiple learning problems, such as supervised learning, reinforcement learning, and self-supervised learning~\citep{springer2025overtrained}---though the authors named the phenomenon over-training. Thus, the generality of plasticity loss has been slowly recognized by the community.

Investigations into plasticity loss have provided insight into the conditions under which the phenomenon is most likely to occur.
\citeauthor{dohare2024loss} (\citeyear{dohare2024loss}) showed, using synthetic experiments based on MNIST, that plasticity loss tends to emerge after extended training on non-stationary data. Moreover, these experiments showed that the onset of the phenomenon occurs earlier in smaller networks. Similarly, \citeauthor{lyle2023understanding} (\citeyear{lyle2023understanding}) showed that larger networks are more robust to the phenomenon in synthetic tasks based on the CIFAR-10 and MNIST datasets. Although insightful, these investigations have been limited to synthetic problems based on vision datasets using feedforward or convolutional networks, and thus provide little evidence for the phenomenon in more realistic models and at scale.

Previous work has also investigated plasticity loss in the language domain, but existing studies leave open whether the phenomenon appears in GPT-style Transformer models trained on realistic natural-language streams for many tasks.
\citeauthor{farias2025selfnormalized} (\citeyear{farias2025selfnormalized}) found severe plasticity loss in a GPT-style model; however, the learning problem used synthetic text data, and the model contained fewer than a million parameters. Similarly, \citeauthor{liu2026neural} (\citeyear{liu2026neural}) found loss of plasticity in a single-layer T5 model~\citep{raffel2020exploring} on a synthetic language problem.
\citeauthor{cho2026forget} (\citeyear{cho2026forget}) presented an algorithm for maintaining plasticity and preventing forgetting when learning from text data using models with more than a billion parameters.
However, their evaluations of plasticity loss included only eight different tasks, which, even in synthetic problems with small networks, may be insufficient to reliably observe the onset of plasticity loss.
Lastly, \citeauthor{springer2025overtrained} (\citeyear{springer2025overtrained}) presented the most comprehensive study of plasticity loss in language models, but focused on the downstream effects of extended pretraining rather than the evolution of the phenomenon in a continual learning setting.

This paper addresses these gaps by studying plasticity loss in continual learning with realistic natural-language data. We study a range of GPT-style Transformer model sizes, complementing the stationary-learning results of \citeauthor{springer2025overtrained} (\citeyear{springer2025overtrained}) with experiments in a non-stationary continual learning regime. Lastly, we train on a large number of tokens and tasks to test whether plasticity loss emerges in natural-language Transformer training and whether its onset changes predictably with model size. Moreover, we believe that our naturalistic multilingual training and evaluation framework provides a useful task to elicit loss of plasticity for future works to build upon. 

\section{Learning Problem}
\label{sec:learning_problem}

Our study of plasticity loss follows the continual learning setting in which a model is trained on a sequence of learning problems, or tasks~\citep{wang2024comprehensive}. Let $\tau \in \{1,2, \dots, k \}$ denote the sequence of tasks encountered during training. Each task $\tau$ has a corresponding space of observations $\mathcal{X}_\tau$, targets $\mathcal{Y}_\tau$, and a probability distribution over observation-target pairs, $(\mathbf{x}, \mathbf{y}) \sim \mathcal{D}_\tau$. The goal of the learning system is to predict the target $\mathbf{y}$ corresponding to an observation $\mathbf{x}$.

We focus on learning systems based on deep neural networks, represented by functions $f_{\boldsymbol{\theta}}$ parameterized by $\boldsymbol{\theta} \in \mathbb{R}^w$, where $w \in \mathbb{N}$. For a task $\tau$, the objective is to find parameters that minimize the expected loss
\begin{equation*}
    \mathcal{L}_\tau (\boldsymbol{\theta}) 
    \overset{.}{=} \mathbb{E}_{(\mathbf{x}, \mathbf{y}) \sim \mathcal{D}_\tau}
    \left[ 
        \ell( f_{\boldsymbol{\theta}}(\mathbf{x}), \mathbf{y})
    \right]
\end{equation*}
where $\ell$ measures the discrepancy between the model's predictions and the target. In practice, we do not have access to the full data distribution. Instead, for each task $\tau$, the model is trained on a finite dataset $D_\tau = {(\mathbf{x}_i, \mathbf{y}_i)}^{N_\tau}_{i=1}$ by minimizing the empirical loss using mini-batch gradient descent.

A fundamental challenge in continual learning is how to incorporate information from new observations without losing previously acquired knowledge, a tension known as the stability-plasticity dilemma~\citep{mermillod2013stability, abel2026plasticity}. Stability refers to the ability to retain performance on previously encountered tasks, while plasticity refers to the ability to learn from new data. Stability has been extensively studied resulting in a large literature dedicated in preventing catastrophic forgetting in neural networks \citep{mccloskey1989, french1999, kirkpatrick2017overcoming}. In this paper, we focus on plasticity: whether a model trained for an extended period retains the ability to improve on a new target distribution under a fixed training budget.

Plasticity can be measured in different ways. One approach is to measure learning online, as the model encounters each new task in the training sequence~\citep{dohare2024loss}. Another approach is to periodically evaluate the model on a held-out probing task that is not part of the training sequence~\citep{lyle2023understanding}. We adopt the latter approach since the former requires tasks to be equally difficult to accurately assess plasticity loss. Thus, at selected points during training, we train on a held-out task and evaluate how efficiently the current model adapts to the new distribution. A degradation in performance over time indicates that the model has lost some of its ability to learn from new observations.

\section{Experimental Setup}
\label{sec:experimental_setup}

\subsection{Multilingual Continual Learning Problem}
\label{subsec:multilingual_continual_learning_problem}

We looked for three desiderata in designing a continual learning problem suitable for studying plasticity loss with large language models:

\begin{enumerate}
\item The data should be natural-language, overcoming the shortcomings of previous studies using synthetic datasets.
\item The sequence of tasks should be extendable indefinitely to enable the training length required to observe measurable effects of plasticity loss.
\item The tasks should be naturalistic and challenging enough for scale to result in meaningful improvements in performance.
\end{enumerate}

Guided by these desiderata, we created a multilingual next-token prediction problem based on sequences of languages from the CulturaX dataset curated by \citeauthor{nguyen2024culturax} (\citeyear{nguyen2024culturax}). The dataset comprises 167 languages and 6.3 trillion tokens.

We selected a subset of eight languages from the CulturaX dataset and then created training and evaluation datasets of 100 billion and 1 billion tokens, respectively, for each selected language. We trained networks on the following sequence of languages: English, written Chinese, French, Japanese, Spanish, German, Portuguese, and Russian, in that order. Each time a language appeared in the sequence, the model was trained on a fresh task instance consisting of 5 billion tokens sampled from the corresponding language's training set. We call a complete sequence of eight task instances a \textit{cycle}, and trained networks for several cycles. We refer to training on this non-stationary sequence of task instances as \textit{continual pretraining}.

\begin{figure}[t]
\centering
\includegraphics[width=\columnwidth]{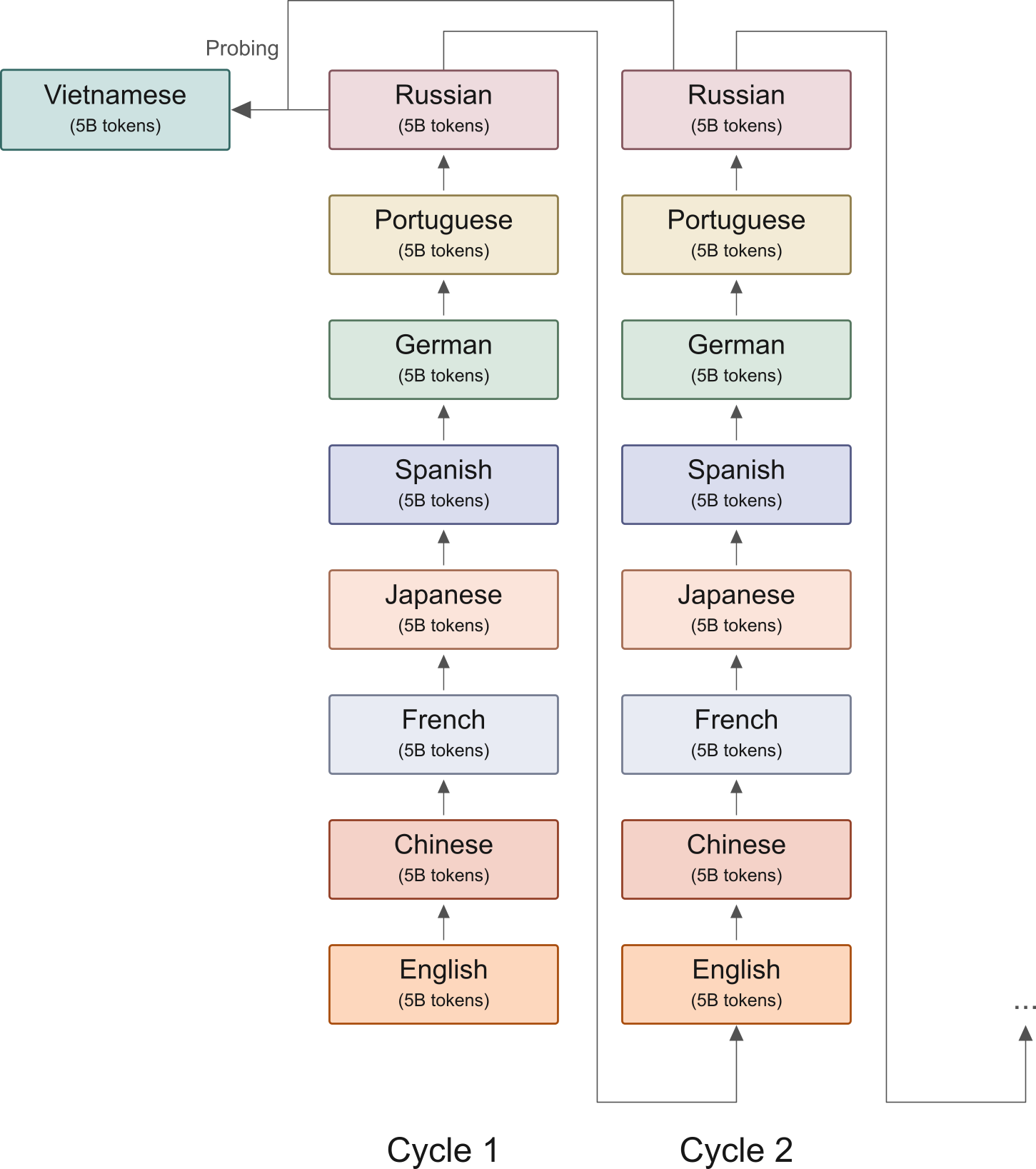}
\caption{Multilingual continual learning problem. Models are trained on repeated cycles of eight language-modeling task instances. At the end of each cycle, plasticity is assessed using a held-out Vietnamese probing task.}
\label{fig:problem_def}
\end{figure}

For the purpose of assessing plasticity loss, we created a separate training and validation set from the Vietnamese language consisting of 20 billion training tokens and 1 billion evaluation tokens, respectively. At the end of each cycle, we conducted a probing task consisting of 5 billion training tokens randomly subsampled from the Vietnamese 20 billion-token training set. Probing was performed on a copy of the model checkpoint, and the resulting parameter updates were discarded before continual pretraining resumed. Thus, Vietnamese data were used only to measure the model's ability to learn new information and did not become part of the continual pretraining sequence. We illustrate the continual pretraining and probing process in Figure~\ref{fig:problem_def}.

The language sequence was selected to reduce similarity between consecutive languages. Similarly, the probing language Vietnamese, as the only austroasiatic language in the training set, was selected to reduce similarity between the probing distribution and the continual pretraining data, limiting transfer and ensuring that probing contained genuinely new information. Since Vietnamese was not included in any cycle, the Vietnamese probing task contained data that had not been observed by model during continual pretraining, and for which we expect transfer from other languages to be low. Consequently, performance by training on the new language indicates how well the model can learn from new observations.

During the probing task, we trained for 5B tokens on the new Vietnamese language data, and measured the validation loss using 1,280 randomly sampled sequences of length 2,048 every 95 training steps. To assess how efficiently the model learned during probing, we used the area under the validation-loss curve as a function of probe-training steps. Lower AUC values indicate that validation loss decreased earlier during probing, corresponding to faster adaptation.

\subsection{Architectures and Training Details}
\label{subsec:architecture_and_training_details}

We trained several GPT-style pre-norm decoder-only Transformer models~\citep{radford2018improving} of varying sizes. All models use causal self-attention, absolute positional embeddings, pre-layer normalization, an MLP expansion factor of 4, GeLU activations, tied input and output embeddings, and a trained embedding table. When choosing the model configurations, we aimed to minimize architectural variation across scales so that changes in performance could be attributed primarily to model size rather than to differences in width-depth trade-offs. We therefore fixed the aspect ratio of all models, defined as \(d_{\mathrm{model}} / L\), to 80, where \(d_{\mathrm{model}}\) is the Transformer hidden dimension and \(L\) is the number of layers. For the same reason, we kept the attention-head dimension fixed at 64, preventing larger models from gaining additional per-head capacity as model size increased.

We trained models at eight different sizes: 5M, 12M, 27M, 39M, 53M, 83M, 106M, and 314M non-embedding parameters. We report model size using non-embedding parameters because the embedding table size is dominated by the tokenizer vocabulary and does not scale in the same way as the Transformer stack. This choice makes model size a better proxy for the capacity of the Transformer layers whose width and depth vary across experiments. For tokenization, we used the pretrained Qwen3 tokenizer~\citep{yang2025qwen3}, which has reasonable coverage across a wide variety of languages and a vocabulary size of 151,680. For total parameter counts and additional architectural details, see Appendix~\ref{app:train_deets}.

All models were trained with a sequence length of 2,048 using AdamW~\citep{loshchilov2018decoupled}, with $\beta_1=0.9$, $\beta_2=0.95$, a weight decay of 0.1, and a batch size of 0.5M tokens. We used a learning rate schedule with linear warm-up for the first 5\% of the total training steps, followed by a constant learning rate. Each task consisted of 9,537 training steps.

We tuned the learning rate using grid searches on the smallest and largest models, training the 5M and 314M models on 5 billion English-language tokens. For each of these two models, we selected the learning rate that minimized the loss under this fixed token budget. The best learning rates for the 5M and 314M models were \(3 \times 10^{-3}\) and \(1 \times 10^{-3}\), respectively. We then interpolated between these two endpoints to determine the learning rates for the intermediate model sizes (see Appendix~\ref{app:train_deets}). To validate this procedure, we compared the losses obtained using the interpolated learning rates with the losses obtained during additional grid searches for intermediate model sizes. The interpolated learning rates produced lower loss than the values tested in those grid searches.

The optimizer was reset at the start of each task; this procedure has been shown to prevent training instabilities caused by sudden changes in the loss landscape~\citep{asadi2023resetting}, and also implies that the plasticity degradations we observed were due to inherent plasticity loss in the weights rather than simply stale optimizer states. To further smooth transitions between tasks, we restarted the learning rate warm-up at the beginning of each task. For further details on hyperparameter selection and values, see Appendix~\ref{app:train_deets}.

\section{Experimental Results}
\label{sec:experimental_results}

\subsection{Emergence and Scaling of Plasticity Loss in Continual Pretraining}
\label{subsec:results_continual_scaling}

At the start of our study, we asked a simple question: \textit{is there evidence of plasticity loss in continual learning problems based on natural-language?} To answer this question, we trained each model on our Multilingual Continual Learning Problem for several cycles and probed at the end of each cycle. We measured learning efficiency using the area under the validation-loss curve (AUC) as the model learned the probing task, with lower AUC indicating faster or more effective adaptation.

\begin{figure*}[t]
\centering
\begin{tikzpicture}
\node[inner sep=0pt] (img) {
\includegraphics[width=0.8\textwidth]{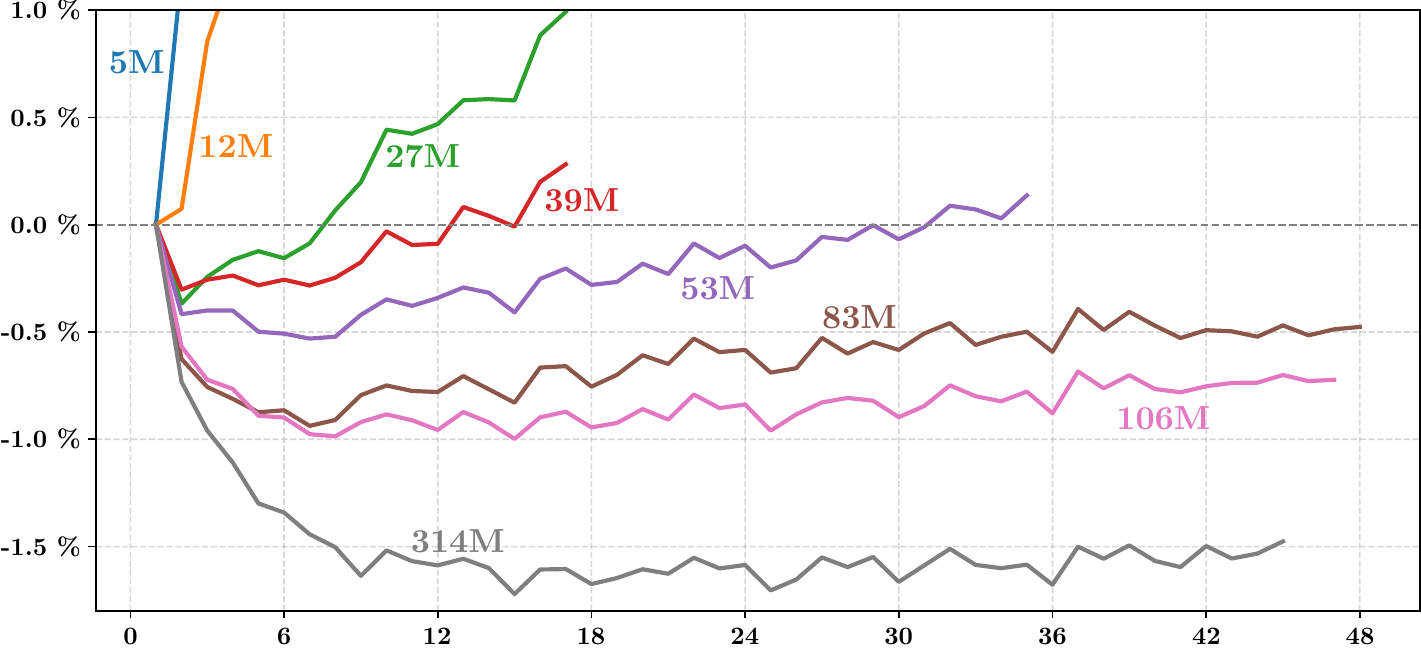}};
\node[anchor=north] at ([yshift=-0.35cm]img.south) {Cycle};
\node[rotate=90, anchor=south] at ([xshift=-0.55cm]img.west) {\% change in AUC relative to first cycle};
\end{tikzpicture}
\caption{Percentage change in validation-loss AUC on the probing task relative to the first cycle for models ranging from 5M to 314M non-embedding parameters from the second to 48th cycle. All the models eventually exhibit an increasing trend in AUC, indicating a reduced ability to learn the probing task and providing evidence of plasticity loss. However for larger models we observe plasticity loss to occur after considerably more cycles. }
\label{fig:lop_all_models}
\end{figure*}

In this setting, a decrease in probing AUC across cycles indicates improved probing performance, consistent with beneficial transfer from continual pretraining to the probing task---i.e. from learning other languages to learning Vietnamese. In contrast, an increasing trend in probing AUC indicates that the model is becoming less able to learn from new data---a loss of plasticity. Since models vary in their absolute performance on Vietnamese due to inherent differences in capacity, we normalize each model's probing performance relative to its performance after the first cycle. Specifically, let $\mathrm{AUC}_k$ denote the AUC measured on the probing task after cycle $k$. For each model, each point in Figure~\ref{fig:lop_all_models} corresponds to

\begin{equation}
100 \times
\frac{\mathrm{AUC}_k - \mathrm{AUC}_1}{\mathrm{AUC}_1} =
100 \times
\left(
\frac{\mathrm{AUC}_k}{\mathrm{AUC}_1} - 1
\right)
\end{equation}

The ratio \(\mathrm{AUC}_k / \mathrm{AUC}_1\) measures the relative change in probing AUC after cycle $k$ with respect to the first cycle, while subtracting one centers the first-cycle performance at zero. Multiplying by 100 expresses this change as a percentage. Thus, positive values indicate higher AUC than in the first cycle, corresponding to worse probing performance, whereas negative values indicate lower AUC and improved probing performance.

All the models show evidence of plasticity loss on the probing task. Models up to 314M parameters eventually show an increasing trend in AUC as the number of cycles increases. Moreover, the onset of this trend occurs earlier in smaller models. For example, the 5M and 12M models show an increasing trend in AUC immediately after the first cycle, whereas the 83M model shows an increasing trend only after the seventh cycle. The results also show that beneficial transfer is more pronounced in larger models: the 53M model achieves a 0.5\% reduction in AUC over time, whereas the 314M model achieves more than a 1.5\% reduction.

We conclude from these results that pre-norm GPT-style Transformer models can lose plasticity when trained on the Multilingual Continual Learning Problem. However, for a fixed training duration, the effect of plasticity loss diminishes with model size: it is more severe in the 5M model than in the 314M model. Moreover, while continual pretraining initially improves performance on the probing language, excessive continual pretraining eventually has a detrimental effect. 

This observation motivates our next research question: \textit{is it possible to predict the onset of plasticity loss based on the size of the model?}; i.e., is it possible to derive a \emph{scaling law} for the onset of the loss of plasticity.

To predict the onset of plasticity loss, we first identified the task after which continual pretraining began to have a detrimental effect on probing performance. In other words, we estimated the minimum of each curve in Figure~\ref{fig:lop_all_models}. Since we probed only once per cycle in the experiments above, we lacked the granularity to precisely identify the task after which the trend began to increase, resulting in multiple models sharing the same minimum. Consequently, we performed additional probing tasks within each cycle.

This higher-frequency probing introduced another issue: each language has a different level of transfer to Vietnamese, which adds noise to the evaluations. For example, pretraining on English consistently resulted in better downstream performance on Vietnamese than pretraining on any other language, whereas pretraining on Japanese consistently resulted in worse performance. These discrepancies due to different levels of transfer often obscured the relatively small effect of plasticity loss. To reduce noise due to different levels of transfer, we probed only after languages with levels of transfer to Vietnamese similar to Russian: Chinese and German. We explain in more detail how we chose these languages in Appendix~\ref{app:predictive_model}. After increasing the probing frequency within each cycle, we further smoothed each curve using a moving average with a window size of three probe measurements. For each model, we identified the minimum of the smoothed curve and used this point as an estimate of the onset of plasticity loss. 


From this data, we then fit a log-log model relating the estimated onset to model size, resulting in the power law
\begin{equation}
T = 1.3 \times 10^{-5} \cdot P^{0.8269}
\label{eq:lop_scaling}
\end{equation}
where $T$ denotes the task-instance number at which we expect to observe measurable effects of plasticity loss and $P$ denotes the number of non-embedding parameters in the model. In Figure~\ref{fig:lop_scaling_model}, we show the fitted line along with the observed data. In Appendix~\ref{app:predictive_model}, we report fits for linear, log-linear, and exponential models. All three alternatives produced higher root mean-squared error than the log-log model under leave-one-out cross-validation, providing some evidence that a power-law fit is most natural.

\begin{figure}[t]
\centering
\begin{tikzpicture}
\node[inner sep=0pt] (img) {
\includegraphics[width=0.8\columnwidth]{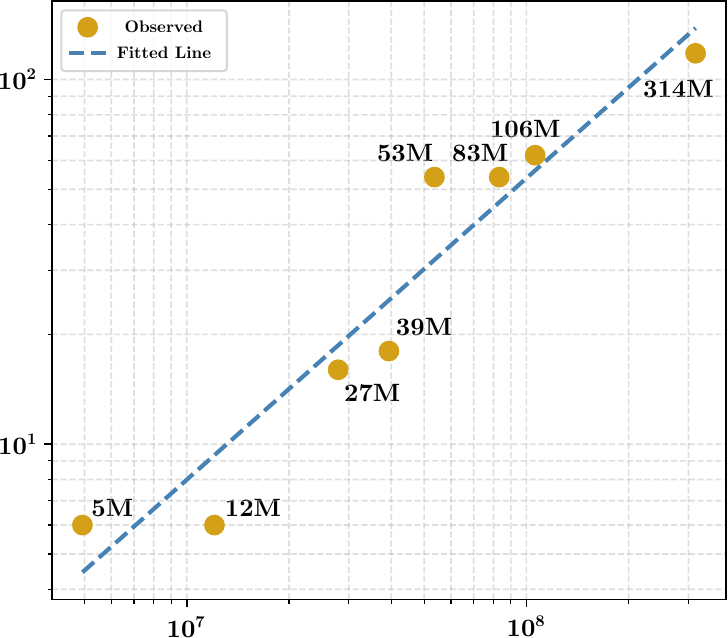}};
\node[anchor=north] at ([yshift=-0.35cm]img.south) {Model size};
\node[rotate=90, anchor=south] at ([xshift=-0.55cm]img.west) {Onset of plasticity loss (task-instance number)};
\end{tikzpicture}
\caption{Log-log model predicting the onset of plasticity loss, measured in number of task instances, as a function of model size. The onset of plasticity loss scales as $T = 1.3 \times 10^{-5} \cdot P^{0.8269}$, where $P$ is the number of non-embedding parameters. We observe a qualitatively decent but not perfect fit with several arguable outliers at 12M and 53M.}
\label{fig:lop_scaling_model}
\end{figure}

The log-log model has an important implication for addressing plasticity loss solely through model scaling. Equation~\ref{eq:lop_scaling} predicts that the number of task instances needed to observe plasticity loss scales sublinearly with the number of parameters. Consequently, increasing model size to prevent plasticity loss has diminishing returns and may be an inefficient approach to addressing the problem.

This scaling relationship also provides a useful point of comparison with recent results on larger language models. \citeauthor{springer2025overtrained} (\citeyear{springer2025overtrained}) provided evidence that the effects of plasticity loss may already be observable in one-billion-parameter models, finding that these models had lower fine-tuning performance after pretraining for longer than 2 trillion tokens. Extrapolating our log-log model to this scale predicts that a model with one billion non-embedding parameters would show signs of plasticity loss after 360 task instances, or 1.8 trillion training tokens, since each task instance consists of 5 billion tokens. Furthermore, they reported no consistent evidence of plasticity loss after training seven-billion-parameter models for 2 trillion tokens. This is also consistent with our extrapolation, which predicts that a model of that size would show evidence of plasticity loss after approximately 9 trillion training tokens.

Although these extrapolations are consistent with the overtraining results reported by \citeauthor{springer2025overtrained} (\citeyear{springer2025overtrained}), the learning problems are substantially different. \citeauthor{springer2025overtrained} (\citeyear{springer2025overtrained}) evaluated models pretrained on stationary datasets, whereas our results so far are based on a non-stationary learning problem with abrupt task changes. This comparison motivates our next question: \textit{does plasticity loss also emerge under stationary training?}

\subsection{Plasticity Loss Under Stationary Training and Network-Level Correlates}
\label{subsec:results_stationary_correlates}

To investigate this question, we trained the 5M, 12M, and 27M models on a stationary mixture of all eight training languages, which we refer to as the Multilingual Stationary Learning Problem. After every 5 billion training tokens, we performed a probing task by training a copy of the current model checkpoint on Vietnamese for 5 billion tokens and measuring the AUC of the validation loss. As in the continual setting, the parameter updates from probing were discarded. We used the same hyperparameters as in the Multilingual Continual Learning Problem, including the number of warm-up steps for the learning rate.

\begin{figure}[t]
\centering
\begin{tikzpicture}
\node[inner sep=0pt] (img) {
\includegraphics[width=0.925\columnwidth]{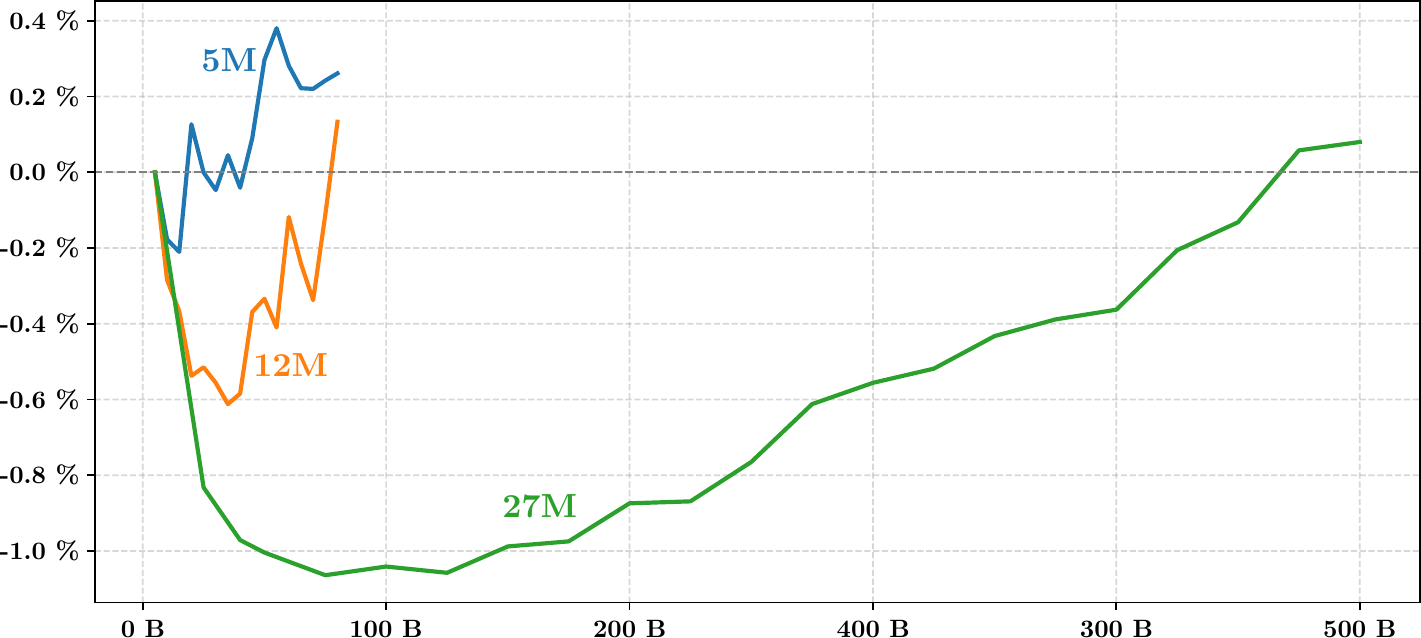}
};
\node[anchor=north] at ([yshift=-0.0cm]img.south) {Training tokens};
\node[rotate=90, anchor=south] at ([xshift=-0.0cm]img.west) {\% change in AUC};
\end{tikzpicture}
\caption{Percentage change in validation-loss AUC on the probing task relative to the first probing task during training on the Multilingual Stationary Learning Problem. As the number of training tokens increases, the AUC of all three models eventually increases, indicating a reduced ability to learn from new information as training progresses.}
\label{fig:lop_stationary}
\end{figure}

The results in the Multilingual Stationary Learning Problem show a pattern similar to that observed in the continual learning setting. Figure~\ref{fig:lop_stationary} shows the percentage change in AUC relative to the first probing task. All three models eventually show an increasing trend in AUC, indicating a decrease in the ability to learn from new data. Moreover, the onset of plasticity loss is, once again, delayed with model size: the 5M model shows plasticity loss sooner than the 27M model. However, the rate at which the models lose plasticity appears to be different. This is most evident in the 27M model, which crosses the zero line after 475B tokens of training in the stationary setting and after 320B tokens in the continual setting. Nevertheless, these results suggest that models trained on stationary data are not exempt from plasticity loss and thus plasticity loss is unlikely to be explained solely by abrupt task changes common in continual learning problems.

Having observed plasticity loss in both continual and stationary learning problems, the next natural question is: \textit{what factors are associated with this phenomenon?} The causes of plasticity loss have been widely debated, and there is currently no clear consensus in the literature ~\citep{lyle2023understanding, lewandowski2024curvature}. Fully characterizing the correlates or causes of plasticity loss is beyond the scope of this paper. Nevertheless, we measure several correlates of plasticity loss that may provide insight into the changes occurring inside the network as its ability to learn deteriorates. Understanding these correlates may also suggest approaches for mitigating plasticity loss.

In this section, we focus on models with 12M, 53M, and 106M non-embedding parameters trained on the Multilingual Continual Learning Problem. Additional measurements for other model sizes are included in Appendix~\ref{app:correlates}.

\begin{figure*}[t]
\centering
\includegraphics[width=\textwidth]{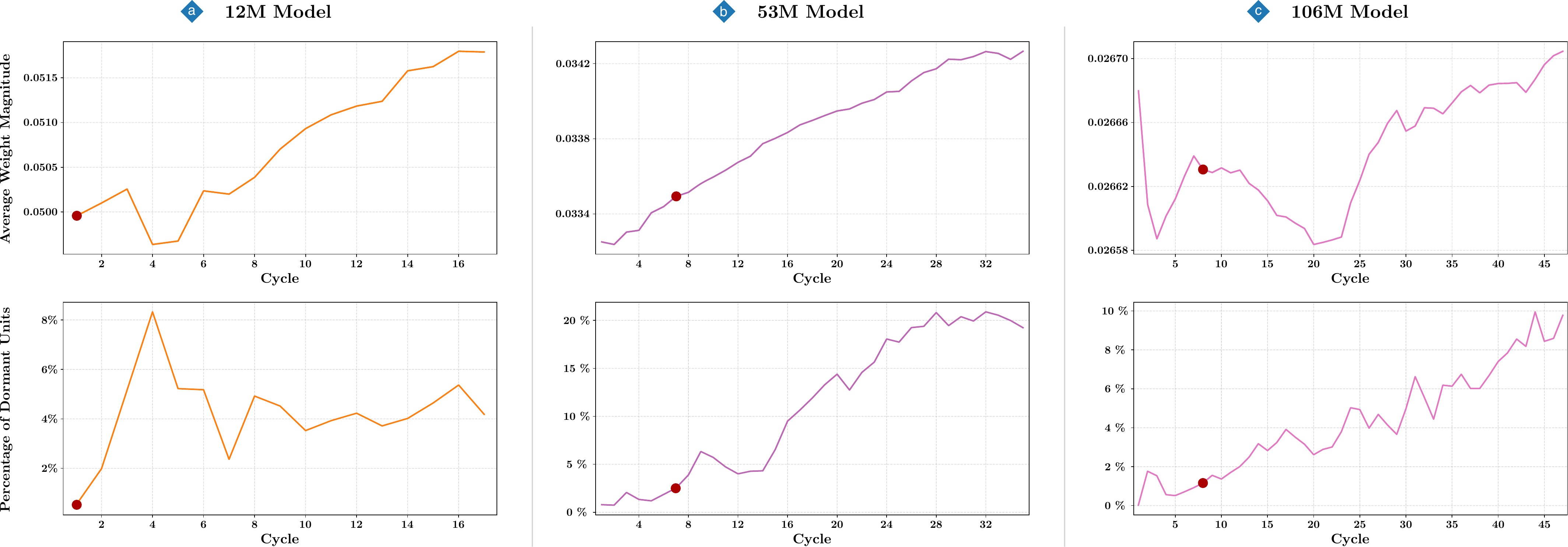}
\caption{Average parameter magnitude and percentage of dormant units for the (a) 12M model, (b) 53M model, and (c) 106M model trained on the Multilingual Continual Learning Problem. As the number of cycles increases, average parameter magnitude and the percentage of dormant units tend to increase across most models. However, these trends do not consistently track the deterioration in probing performance due to plasticity loss. The red marker indicates the cycle after which the model's probing performance began to deteriorate.}
\label{fig:avg_mag_dorm}
\end{figure*}

We recorded metrics previously observed to correlate with plasticity loss. First, we measured the average absolute magnitude of the model parameters, excluding embedding layers. The average is computed over all non-embedding parameters in the network, including both weights and biases; it is not computed separately per layer before averaging. Average parameter magnitude has repeatedly been observed to correlate with plasticity loss~\citep{kumar2024maintaining, dohare2024loss, hernandez2025reinitializing}, making it a useful diagnostic for the deterioration in probing performance observed in Figure~\ref{fig:lop_all_models}. The top row of Figure~\ref{fig:avg_mag_dorm} shows the average parameter magnitude measured at the end of each cycle for the 12M, 53M, and 106M models. The red marker in each plot corresponds to the cycle after which the model's probing performance began to deteriorate.

We found that average parameter magnitude tended to increase with the number of cycles. However, this pattern did not consistently track the observed deterioration in probing performance. For example, although the 106M model starts showing deterioration in performance on the eighth cycle (Figure~\ref{fig:avg_mag_dorm}c, top row), its average parameter magnitude decreases steadily between the eighth and twentieth cycles. Conversely, for the 53M model, probing performance improves between the first and seventh cycles, even though average parameter magnitude steadily increases during the same period (Figure~\ref{fig:avg_mag_dorm}b, top row). Thus, it appears that the average parameter magnitude alone does not fully account for the plasticity loss observed in these models.

To further investigate correlates of plasticity loss, we measured the percentage of dormant units in the network's MLP layers. We follow the definition of dormant units from \citet{sokar2023the}, adapting it to the sequence-modeling setting. Let $h_i^l(\mathbf{x}, j)$ denote the activation of MLP unit $i$ in layer $l$ at token position $j$ for sequence $\mathbf{x}$. Given a dataset $D$ of sequences of length $m$, let the average absolute activation of this unit be
\begin{equation}
\bar{h}^l_i =
\frac{1}{\lvert D \rvert \cdot m}
\sum_{\mathbf{x} \in D}
\sum_{j=1}^{m}
\left| h^l_i(\mathbf{x}, j) \right|
\end{equation}
We then define the normalized absolute activation as
\begin{equation}
s_i^l = \frac{\bar{h}^l_i}{\frac{1}{N^l} \sum_{k = 1}^{N^l} \bar{h}^l_k}
\end{equation}
where $N^l$ is the number of units in layer $l$. A unit $i$ in layer $l$ is said to be $\epsilon$-dormant if $s^l_i \leq \epsilon$, for some small positive threshold $\epsilon$. A high fraction of dormant units suggests that a large portion of the layer has low activity relative to other units in the same layer. In networks with smooth activations such as GeLU, dormancy should not be interpreted as exact inactivity, but rather as a diagnostic indicating that some units contribute comparatively little to the layer's representation.

In our experiments, we measured the percentage of dormant units in the network at the end of each cycle. We measured average absolute activations on 256 sequences of length 2,048 sampled from the Vietnamese validation dataset, which the model had not observed during continual pretraining. We set the threshold to $\epsilon=0.01$. The bottom row of Figure~\ref{fig:avg_mag_dorm} shows the results for the 12M, 53M, and 106M models.

\begin{figure}[t]
\centering
\includegraphics[width=\columnwidth]{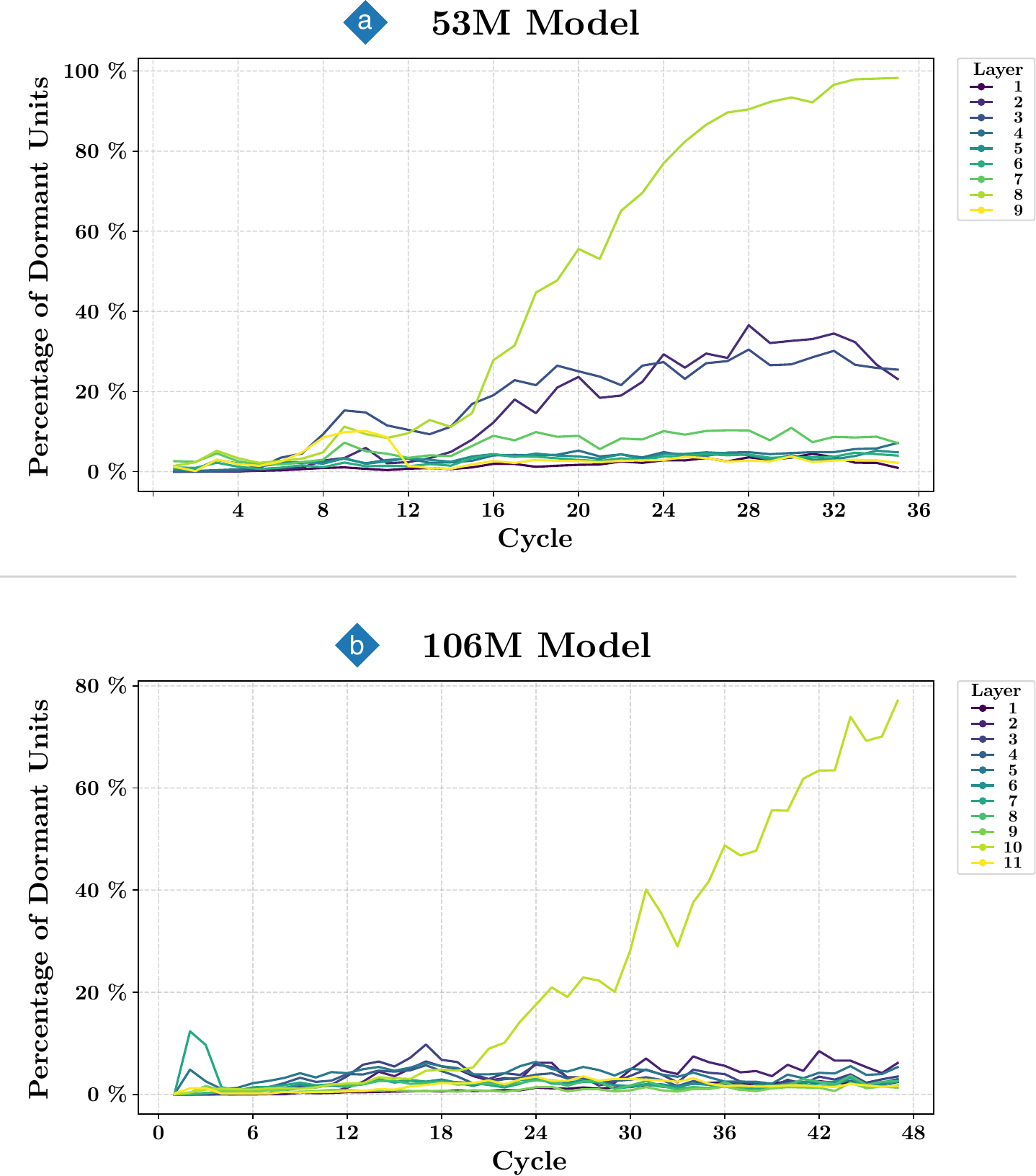}
\caption{Percentage of dormant units per layer in the (a) 53M and (b) 106M models trained on the Multilingual Continual Learning Problem. In both models, one layer accumulates a disproportionately large fraction of dormant units, suggesting that many units in that layer have low activity relative to other units in the same layer.}
\label{fig:dorm_per_layer}
\end{figure}

In the 53M and 106M models, we observed an increasing trend in the percentage of dormant units in the network. These dormant units were not evenly distributed across layers. In the 53M model, more than 95\% of the units in the eighth layer were dormant (Figure~\ref{fig:dorm_per_layer}a). Similarly, in the 106M model, nearly 80\% of the units in the tenth layer were dormant (Figure~\ref{fig:dorm_per_layer}b). Nevertheless, dormant units alone cannot fully explain plasticity loss, since we did not observe an increasing percentage of dormant units in the 12M model (bottom row of Figure~\ref{fig:avg_mag_dorm}a) from cycle 5 onward, even though this model exhibited the most severe plasticity loss and the performance continued to deteriorate after cycle 5.

Having observed increasing dormancy in MLP layers over the course of training, we turned our attention to the attention layers. Specifically, we measured the entropy of attention heads in the network. For a single query position, let $\mathbf{p} \in \mathbb{R}^s$ denote the normalized attention weights over $s$ keys. The entropy of this attention distribution is defined as
\begin{equation*}
\text{Entropy}(\mathbf{p}) = - \sum^s_{i=1} \mathbf{p}_i \log (\mathbf{p}_i).
\end{equation*}
In attention heads, entropy measures how attention is distributed across elements in the sequence. High entropy corresponds to a diffuse distribution of attention weights, assigning nearly equal weight to many elements. High-entropy attention can indicate underutilized attention heads, which may be detrimental to model performance~\citep{sanyal2026when}. Intuitively, when attention is nearly uniform over a long sequence, the head primarily averages information across many positions, limiting its ability to provide query-specific information. Conversely, low entropy indicates that attention is concentrated on a small number of elements in the sequence. This phenomenon, known as entropy collapse, has been linked to training instabilities in Transformer architectures~\citep{zhai2023stabilizing}. One possible contributor to low-entropy attention is the emergence of attention sinks~\citep{xiao2024streamingllm,gu2025attention_sink}, in which certain tokens, often special tokens such as the \texttt{BOS} token, receive a disproportionately large amount of attention mass.

We measured the entropy of attention heads at the end of each cycle using the Russian-language validation dataset. For each cycle, we sampled 512 sequences of length 2,048 and computed the attention distribution for query positions 256 through 2,048. We then computed the entropy of each attention distribution and averaged across sequences and query positions to obtain the average entropy per head, per layer, and per cycle. The maximum achievable average entropy over these query positions is
\begin{equation}
H_{\max} = \frac{1}{1793} \sum^{2048}_{k=256} \ln(k)
\approx 6.9215
\end{equation}
We categorized heads as \textit{lazy} if their average entropy was above $\frac{9}{10} \cdot H_{\max}$, indicating attention patterns that are close to uniform across most available context positions. Conversely, we categorized heads as \textit{collapsed} if their average entropy was below $\frac{1}{10} \cdot H_{\max}$, indicating attention patterns concentrated on very few positions.

\begin{figure*}[t]
\centering
\includegraphics[width=\textwidth]{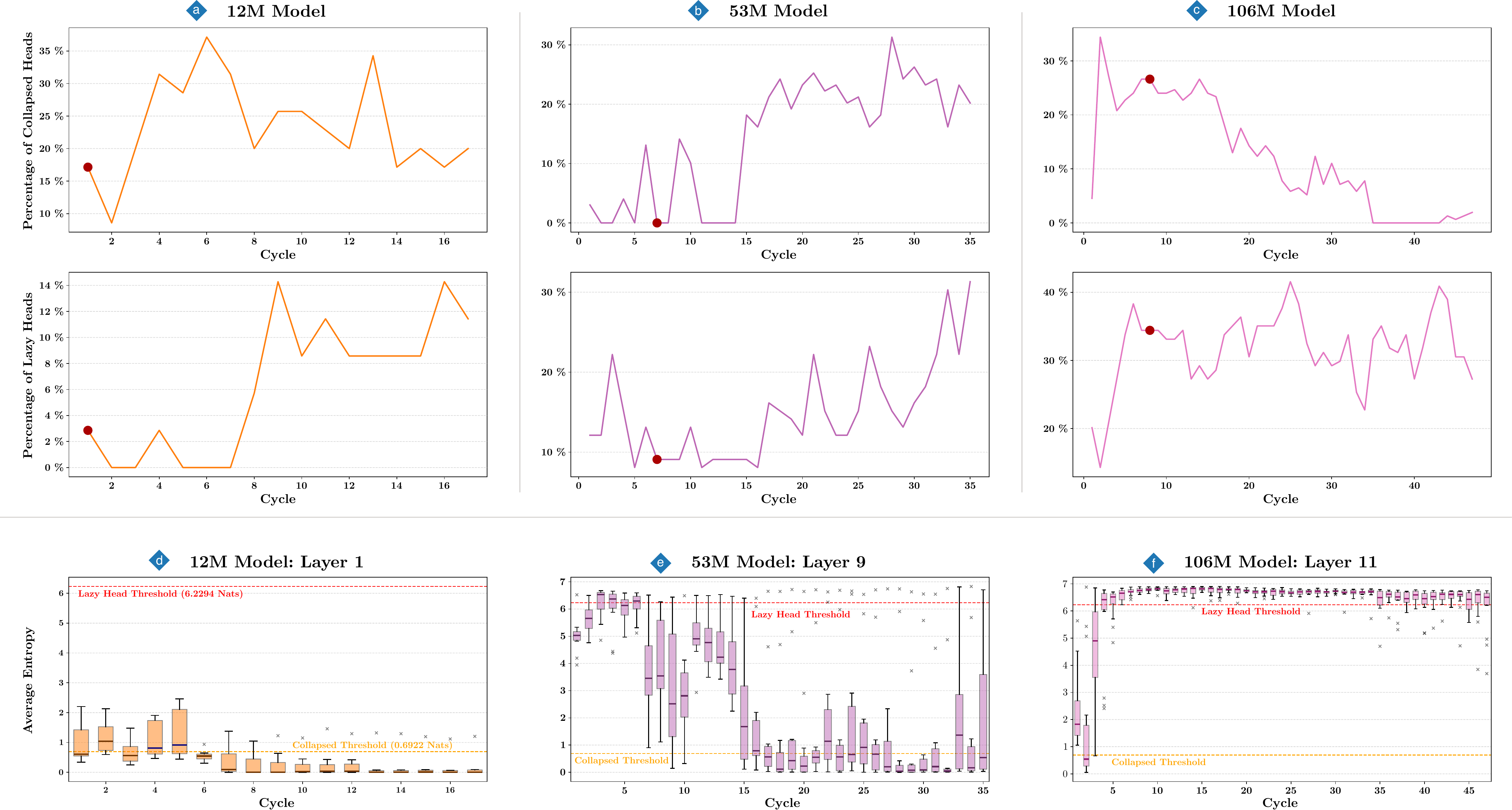}
\caption{Top panels: Percentage of collapsed and lazy attention heads for the (a) 12M model, (b) 53M model, and (c) 106M model trained on the Multilingual Continual Learning Problem. Bottom panels: entropy of attention heads for (d) the first layer of the 12M model, (e) the ninth layer of the 53M model, and (f) the eleventh layer of the 106M model. Neither the percentage of collapsed heads nor the percentage of lazy heads fully explains plasticity loss. Nevertheless, these metrics reveal pathological changes in the attention layers. In extreme cases, entire layers collapse or become lazy.}
\label{fig:attn_entropy}
\end{figure*}

We report the percentage of lazy and collapsed heads per cycle for the 12M, 53M, and 106M models in Figure~\ref{fig:attn_entropy}. The red markers correspond to the cycle after which probing performance started to deteriorate. As before, the percentage of lazy or collapsed heads does not perfectly correlate with plasticity loss. For instance, in the 53M model, the percentage of collapsed heads increases with the number of cycles (top row of Figure~\ref{fig:attn_entropy}b), whereas it decreases for the 106M model (top row of Figure~\ref{fig:attn_entropy}c). For the 106M model, the increase in the percentage of lazy heads precedes the onset of plasticity loss (bottom row of Figure~\ref{fig:attn_entropy}c), whereas in the 12M model the increase occurs after the onset of plasticity loss (bottom row of Figure~\ref{fig:attn_entropy}a). In one case, the percentage of collapsed heads follows the opposite trend from the deterioration in probing performance (top row of Figure~\ref{fig:attn_entropy}c).

Neither the percentage of collapsed heads nor the percentage of lazy heads fully explains plasticity loss. Nevertheless, these metrics point to pathological changes occurring inside the network. In some cases, they reveal layer-wide changes in attention behavior. In the 12M model, the first layer increasingly collapses as the number of cycles increases (Figure~\ref{fig:attn_entropy}d). In the 53M model, the ninth layer collapses for several cycles, though it later recovers (Figure~\ref{fig:attn_entropy}e). Lastly, the eleventh layer of the 106M model becomes lazy after the first few cycles (Figure~\ref{fig:attn_entropy}f). These examples suggest that entire attention layers can enter regimes in which their attention patterns become either overly concentrated or overly diffuse, potentially reducing their contribution to the model's performance.

Although the main text focuses on three model sizes, the additional measurements reported in Appendix~\ref{app:correlates} show a similar qualitative picture. None of the three metrics perfectly tracks the onset or severity of plasticity loss across all models. However, taken together, they suggest that continued training changes the networks in ways consistent with reduced effective capacity and slower adaptation. In the MLP layers, this appears as an accumulation of low-activity units. In the attention layers, it appears as an increase in heads whose attention patterns become either overly diffuse or overly concentrated. In parallel, average parameter magnitude tends to increase, which, in the presence of layer normalization, may contribute to slower learning, as pointed out by \citeauthor{lyle2024normalization} (\citeyear{lyle2024normalization}). Together, these effects may contribute to the reduced ability to learn from new data.

This account is necessarily partial. Our measurements are correlational and do not provide a mechanistic explanation for why low-activity units, lazy attention heads, collapsed attention heads, or increased parameter magnitude emerge over training. Nevertheless, these diagnostics suggest possible directions for designing learning systems that maintain plasticity over longer training horizons.

One simple approach is to limit parameter growth. In our experiments, we used a weight decay of 0.1, a value commonly used in language-model training. However, recent work by \citeauthor{han2026weight} (\citeyear{han2026weight}) has shown that higher values of weight decay can improve model plasticity despite resulting in higher pretraining loss. Relatedly, previous work has suggested clipping parameter values near zero as an effective way to prevent plasticity loss~\citep{elsayed2024weight}. Both approaches directly target the increase in average parameter magnitude observed in the top row of Figure~\ref{fig:avg_mag_dorm}.

A second approach is to mitigate potential capacity loss caused by the accumulation of dormant units. For example, Continual Backpropagation measures the activity of units and periodically reinitializes units with low average activity~\citep{dohare2021continual, dohare2024loss}. Other algorithms, such as ReDo~\citep{sokar2023the}, Self-Normalized Resets~\citep{farias2025selfnormalized}, and GraMa~\citep{liu2026measure}, also reinitialize neurons but use different criteria for selecting which units to reset. Alternatively, one can design activation functions that are less prone to producing dormant units~\citep{abbas2023loss, lillo2026activation}.

The increase in lazy or collapsed attention heads has not been directly studied in the plasticity-loss literature. Nevertheless, existing methods suggest possible interventions. Shrink-and-Perturb~\citep{ash2020on} and Utility-based Perturbed Gradient Descent~\citep{elsayed2024addressing}, for example, add noise to network parameters and may help move attention heads out of overly diffuse or overly concentrated regimes. Similarly, Selective Weight Reinitialization operates directly on the network's weights rather than on individual units~\citep{hernandez2025reinitializing}, and may help counteract the accumulation of pathological attention patterns. For overly diffuse attention, previous work has proposed adding learnable tokens that act as attention sinks, collecting attention mass without affecting the predictions of real sequence elements~\citep{xiao2024efficient, darcet2024vision}. Whether these methods can prevent the attention-pathology patterns observed here remains an open empirical question.

\section{Conclusion}
\label{sec:conclusion}

In this paper, we studied loss of plasticity in GPT-style Transformer models trained on natural-language data. Whereas previous studies of plasticity loss have largely focused on smaller models, synthetic problems, or non-language domains, our results show that GPT-style models with more than 100 million non-embedding parameters can lose the ability to efficiently adapt to new data even when trained on real, large-scale natural-language datasets.

Beyond demonstrating plasticity loss in this setting, we are the first to fit a model predicting the onset of plasticity loss as a function of model size. To make this comparison more controlled, we kept the aspect ratio and attention-head dimension fixed across model sizes, reducing architectural variation beyond scale. The resulting model is a classical power-law `scaling-law' between the number of parameters and the number of tokens until the onset of plasticity loss. Our law predicts that plasticity loss scales sublinearly with the number of non-embedding parameters, implying diminishing returns from increasing model size alone as a strategy for preventing plasticity loss. This prediction speaks to the question posed in the title of this paper: scale alone cannot save us from plasticity loss.

Our result is even more striking given that we also observed evidence of plasticity loss under stationary multilingual training. This means that plasticity loss is not simply caused by our original, somewhat artificially nonstationary task where we cycle through languages. Rather, plasticity loss also seems to affect models which have undergone regular pretraining for a long period on a stationary dataset. Overall, then, we can tentatively conclude that scaling will delay plasticity loss, often for a substantial number of tokens, but cannot fully defeat it, and secondly that we should expect plasticity loss to eventually affect standard language models which have been trained using standard methods. This implies that loss of plasticity is not an artifact of small networks, outdated architectures, or unusual training setups, but rather a fundamental and extremely general property of neural network training. 

Lastly, we measured several correlates of plasticity loss inside the network. As models lost plasticity, they tended to accumulate low-activity MLP units, lazy attention heads, collapsed attention heads, and larger parameter magnitudes. These effects did not appear uniformly across all models, and none of them alone fully explains the observed deterioration in probing performance. Nevertheless, they provide useful diagnostics and suggest possible directions for mitigation, ranging from stronger constraints on parameter growth to methods that track unit activity or reinitialize underused components. We hope these results help guide the design of learning systems that maintain plasticity over long training horizons in both stationary and continual learning settings.

\bibliographystyle{tmlr}
\bibliography{main}
\clearpage

\appendices

\section{Further Details on Parameter Counts, Architectural Choices, and Hyperparameter Selection}
\label{app:train_deets}

Table~\ref{tab:model-configs} reports the architectural configurations and parameter counts for the models used in our experiments. Across model sizes, we kept the aspect ratio \(d/L\), where \(d\) is the hidden dimension and \(L\) is the number of layers, as close to 80 as possible. We also fixed the attention-head dimension, \(d/\text{Heads}\), to 64 for all models. The name of each model refers to the number of non-embedding parameters.

\begin{table*}[h]
\centering
{\small
\setlength{\tabcolsep}{4pt}
\begin{tabular}{lrrrrrrr}
\toprule
\textbf{Model} & \textbf{Layers} & \textbf{Hidden} & \textbf{Attn.} & \textbf{\(d/L\)} & \textbf{\(d/\mathrm{Heads}\)} & \textbf{Eff. Params} & \textbf{Total Params} \\
& \textbf{\(L\)} & \textbf{Dim. \(d\)} & \textbf{Heads} & & & & \\
\midrule
5M   & 4  & 320  & 5  & 80.0 & 64 & 4,932,480   & 54,125,440  \\
12M  & 5  & 448  & 7  & 89.6 & 64 & 12,072,256  & 80,942,400  \\
27M  & 7  & 576  & 9  & 82.3 & 64 & 27,922,752  & 116,470,080 \\
39M  & 8  & 640  & 10 & 80.0 & 64 & 39,389,440  & 137,775,360 \\
53M  & 9  & 704  & 11 & 78.2 & 64 & 53,610,304  & 161,834,816 \\
83M  & 10 & 832  & 13 & 83.2 & 64 & 83,176,704  & 211,078,400 \\
106M & 11 & 896  & 14 & 81.5 & 64 & 106,101,632 & 243,841,920 \\
314M & 16 & 1280 & 20 & 80.0 & 64 & 314,841,600 & 511,613,440 \\
\bottomrule
\end{tabular}
}
\caption{Model configurations used in our experiments. All models use an attention-head dimension, \(d/\mathrm{Heads}\), of 64. The aspect ratio, \(d/L\), was kept as close to 80 as possible. Eff. Params excludes embedding layers, whereas Total Params includes all parameters in the network.}
\label{tab:model-configs}
\end{table*}

\begin{figure}[t]
\centering
\includegraphics[width=0.5\textwidth]{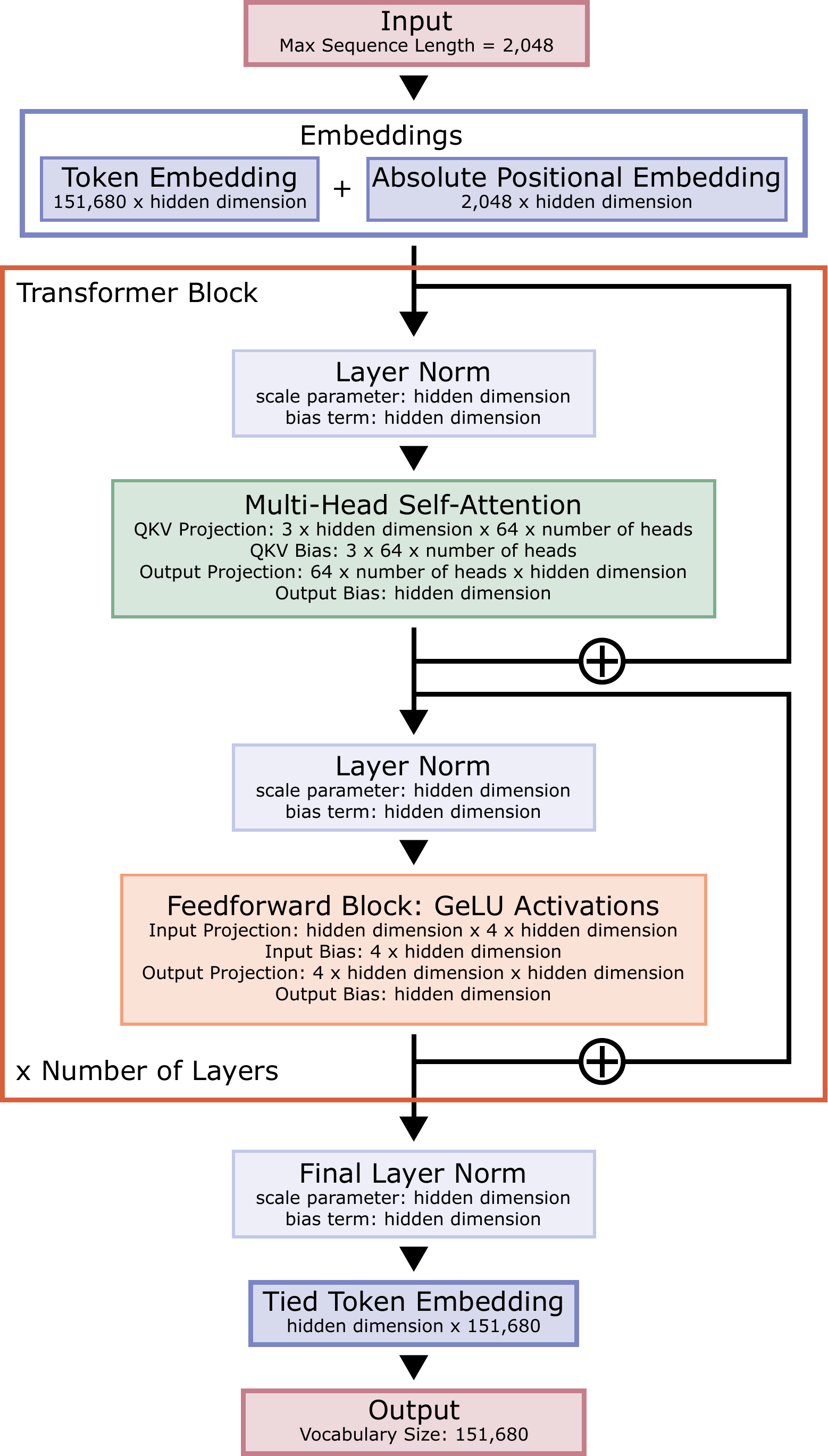}
\caption{Diagram of the specific Transformer architecture used in our experiments.}
\label{fig:arch}
\end{figure}

Figure~\ref{fig:arch} shows the specific Transformer architecture used in our study of plasticity loss. Notably, the architecture uses absolute positional encoding with tied embeddings, layer normalization before the attention and feedforward blocks (pre-norm), and GeLU activations. 

The learning rate, \(\eta\), was tuned for the 5M and 314M models using a grid search over final validation loss after 5 billion tokens of training on English. We tested values of
\begin{align}
\eta \in &\left\{
1 \times 10^{-4},
3 \times 10^{-4},
5 \times 10^{-4},
8 \times 10^{-4},\right.\\
&\left.1 \times 10^{-3},
3 \times 10^{-3},
5 \times 10^{-3}
\right\}
\end{align}
The best learning rates were \(3 \times 10^{-3}\) for the 5M model and \(1 \times 10^{-3}\) for the 314M model. We then interpolated learning rates for the intermediate-size models by modeling the learning rate as a power law in the hidden dimension, similar in spirit to the scaling rule used in \(\mu\)P:

\begin{align*}
\eta_{\text{314M}}
&= \eta_{\text{5M}}
\left(
\frac{d_{\text{5M}}}{d_{\text{314M}}}
\right)^\alpha, \\
\alpha
&= \frac{\log(\eta_{\text{314M}} / \eta_{\text{5M}})}
{\log(d_{\text{5M}} / d_{\text{314M}})}
= \frac{\log(1/3)}{\log(1/4)}
\approx 0.79248.
\end{align*}

We verified that this interpolation was sensible by comparing the interpolated learning rates with additional grid-search results for the 12M, 27M, 53M, and 106M models. In every case, the interpolated learning rate yielded a lower final validation loss on English than the other grid-search values tested. The learning rates obtained from this interpolation, together with the grid-search results for each model, are listed in Table~\ref{tab:learning-rates}.

\begin{table}[t]
\centering
{\small
\setlength{\tabcolsep}{4pt}
\begin{tabular}{lccccc}
\toprule
\textbf{Model} & \textbf{\(d\)} & \multicolumn{2}{c}{\textbf{Interpolation}} & \multicolumn{2}{c}{\textbf{Grid Search}} \\
\cmidrule(lr){3-4} \cmidrule(lr){5-6}
& & \(\eta \times 10^{3}\) & Loss & \(\eta \times 10^{3}\) & Loss \\
\midrule
5M   & 320  & 3.00 & 3.9941 & 3.00 & 3.9941 \\
12M  & 448  & 2.30 & 3.7973 & 3.00 & 3.8088 \\
27M  & 576  & 1.88 & 3.6303 & 1.00 & 3.6394 \\
39M  & 640  & 1.73 & 3.5767 & --   & --     \\
53M  & 704  & 1.61 & 3.5100 & 1.00 & 3.5130 \\
83M  & 832  & 1.41 & 3.4436 & --   & --     \\
106M & 896  & 1.33 & 3.4006 & 1.00 & 3.4037 \\
314M & 1280 & 1.00 & 3.2575 & 1.00 & 3.2575 \\
\bottomrule
\end{tabular}
}
\caption{Learning rates per model size, scaled as \(\eta = \eta_{\mathrm{5M}} \cdot (d_{\mathrm{5M}}/d)^{0.79248}\). The loss column was computed by averaging the last 10 validation-loss measurements. Each validation-loss measurement was computed using 2.5 million tokens.}
\label{tab:learning-rates}
\end{table}

\section{Finding the Minimum of the Plasticity Loss Curve and Creating a Predictive Model}
\label{app:predictive_model}

To create the predictive model presented in the main text, we first identified, for each model, the task after which the trend in Figure~\ref{fig:lop_all_models} began to increase. Initially, we only probed after Russian, the last language in the cycle. This approach was ideal for demonstrating plasticity loss because any deterioration in performance could be attributed solely to a loss of learning ability. However, this probing approach provided a coarse level of granularity, preventing us from more precisely determining the start of the increasing trend in performance. 

To increase the granularity in our probing, we decided to perform more probing tasks within each cycle. However, probing after each language resulted in noisy data because performance on Vietnamese was heavily influenced by the preceding language. To illustrate this, we show in Table \ref{tab:transfer} the AUC of the validation loss on the Vietnamese language after pretraining, individually, on each of the eight different languages in the cycle. We found that some languages consistently resulted in better transfer than others. For example, pretraining on English consistently resulted in the lowest AUC across the 5M, 12M, and 27M models. On the other hand, pretraining on Japanese consistently resulted in higher AUC than the average across languages. 

To reduce noise from the pretraining language preceding probing, we selected languages with similar levels of transfer. Thus, we selected Chinese, German, and Russian for extended probing, corresponding to the second, sixth, and eighth language in a cycle, respectively. We also considered including Japanese, but the difference in transfer between Japanese and German once again introduced too much noise.

\begin{table}[t]
\centering
{\small
\setlength{\tabcolsep}{4pt}
\begin{tabular}{lccc}
\toprule
\multirow{2}{*}{\makecell{\textbf{Pretraining} \\ \textbf{Language}}} 
& \multicolumn{3}{c}{\textbf{AUC on Vietnamese Language}} \\
\cmidrule(lr){2-4}
  & \textbf{5M Model} & \textbf{12M Model} & \textbf{27M Model} \\
\midrule
English    & 3.0288  & 2.8189 & 2.6496 \\
\rowcolor{blue!10} 
Chinese    & 3.0634  & 2.8602 & 2.6925 \\
French     & 3.0498  & 2.8440 & 2.6718 \\
Japanese   & 3.0661  & 2.8746 & 2.7019 \\
Spanish    & 3.0502  & 2.8423 & 2.6710 \\
\rowcolor{blue!10} 
German     & 3.0560  & 2.8519 & 2.6842 \\
Portuguese & 3.0562  & 2.8506 & 2.6712 \\
\rowcolor{blue!10} 
Russian    & 3.0793  & 2.8641 & 2.6932 \\
\midrule
Average    & 3.0562  & 2.8508 & 2.6794 \\
\bottomrule
\end{tabular}
}
\caption{Performance on Vietnamese after pretraining on different languages for the 5M, 12M, and 27M models. The area under the curve (AUC) of the validation loss measured while training on Vietnamese varies depending on the pretraining language. Pretraining on English consistently results in better transfer than pretraining on any other language. Highlighted rows correspond to the languages selected for extended probing.}
\label{tab:transfer}
\end{table}

\begin{figure*}[t]
    \centering
    \includegraphics[width=\textwidth]{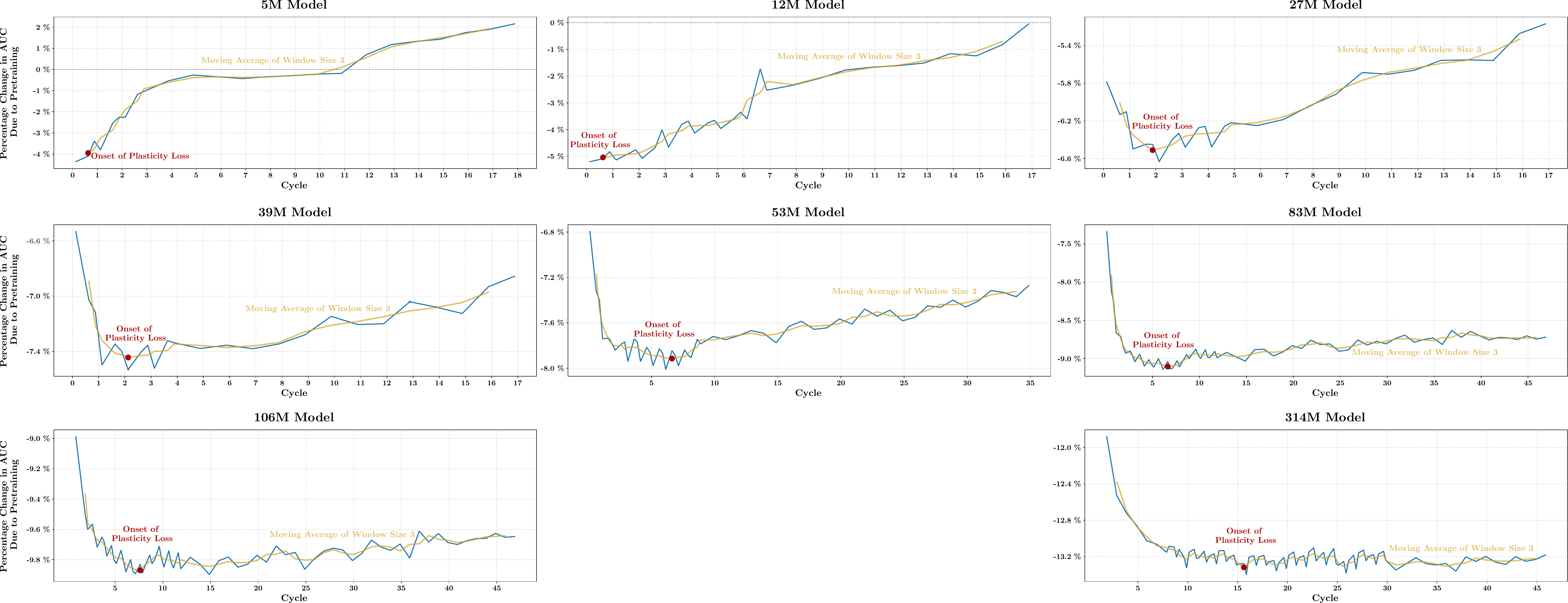}
    \caption{Extended probing in the Multilingual Continual Learning Problem. Each point is the AUC of the validation loss during the probing task divided by the AUC when training on the Vietnamese language alone. The yellow line corresponds to a moving average with a window size of 3. The red dot corresponds to the onset of plasticity loss used for building predictive models as a function of model size. The onset of plasticity loss occurs later with increasing model size.}
    \label{fig:plast_smooth}
\end{figure*}

After extensive probing, we normalized the AUC line by dividing it by the area under the validation loss curve when training on Vietnamese alone. The blue lines in the plots in Figure \ref{fig:plast_smooth} correspond to the change in AUC relative to training on Vietnamese alone, i.e., the change in AUC due to continual pretraining. Note that we only extended probing in cycles near the minimum of the line to minimize computational cost. 

To further account for the effect of the pretraining language immediately before probing, we computed a moving average with a window size of three. Lastly, we used the minimum point of the moving average line as the onset of plasticity loss. The yellow lines in the plots in Figure \ref{fig:plast_smooth} correspond to the moving average line with a point indicating the minimum point. Table \ref{tab:plasticity-onset} shows the task after which the trend in AUC increased---the onset of plasticity loss---along with the parameter count of the model. 

\begin{table}[t]
\centering
{ \small
\begin{tabular}{lcc}
\toprule
\textbf{Model} & \makecell{\textbf{Effective Size} \\ $P$} & \makecell{\textbf{Onset of Plasticity Loss} \\ $T$} \\
\midrule
5M   & 4,932,480   & 6   \\
12M  & 12,072,256  & 6   \\
27M  & 27,922,752  & 16  \\
39M  & 39,389,440  & 18  \\
53M  & 53,610,304  & 54  \\
83M  & 83,176,704  & 54  \\
106M & 106,101,632 & 62  \\
314M & 314,841,600 & 118 \\
\bottomrule
\end{tabular}}
\caption{Onset of plasticity loss, in number of tasks, for each model size. The onset was obtained by taking the minimum of the change in AUC plots after applying a moving average of window size 3.}
\label{tab:plasticity-onset}
\end{table}

We used the data in Table \ref{tab:plasticity-onset} to fit four models for predicting the onset of plasticity loss, each encoding a different belief about how the onset, $T$, scales with model size, $P$. To encode the belief that onset increases by a constant amount per parameter, we fitted a linear model: $T = a + b \cdot P$. To encode the belief that the onset increases by a constant amount per order-of-magnitude increase in parameters, we fitted a linear-log model: $T = a + b \cdot \ln P$.
We fitted the exponential model: $T = a \cdot e^{b \cdot P}$, which encodes the belief that each additional parameter has a compounding effect on the onset.

Finally, we fitted the log-log model: $\ln T = a + b \cdot \ln P$, which suggests the onset grows multiplicatively with model size, i.e., multiplying $P$ by a scalar changes $T$ by a constant factor. 

\begin{figure*}[t]
    \centering
    \includegraphics[width=\textwidth]{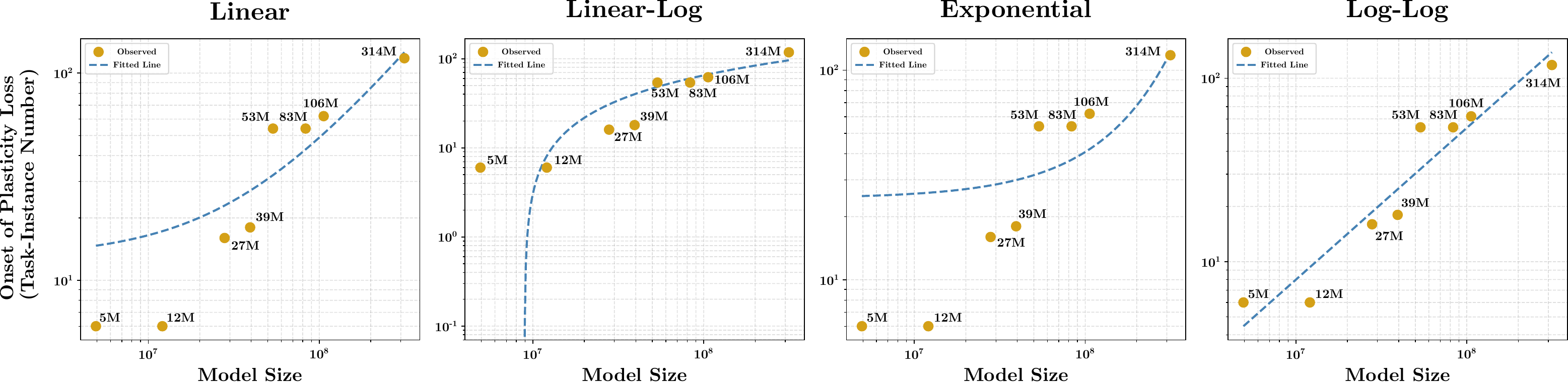}
    \caption{
    Comparison between different models for predicting the onset of plasticity loss. 
    }
    \label{fig:model_comp}
\end{figure*}

We compared the four models using the $R^2$ value, the root-mean-squared error computed on the data (in-sample RMSE), and the RMSE computed using a leave-one-out cross-validation (out-of-sample RMSE).
The coefficients and comparison metrics are given in Table \ref{tab:model-comparison}. Figure \ref{fig:model_comp} shows the fitted line of each model.  Of all the models, the log-log models scored the best.

\begin{table*}[t]
\centering
{\small
\begin{tabular}{lccccccc}
\toprule
\multicolumn{2}{c}{\textbf{Model}} & \multicolumn{3}{c}{\textbf{Coefficients}} & \multirow{2}{*}{$R^2$} & \multirow{2}{*}{\makecell{\textbf{In-Sample} \\ \textbf{RMSE}}} & \multirow{2}{*}{\makecell{\textbf{Out-of-Sample} \\ \textbf{RMSE}}} \\
\cmidrule(lr){1-2} \cmidrule(lr){3-5}
Name & Formula & Name & Value & STE & & & \\
\midrule
\multirow{2}{*}{Linear} & \multirow{2}{*}{$T = a + b \cdot P$} & $a$ & $12.9086$ & $6.3401$ & \multirow{2}{*}{0.8914} & \multirow{2}{*}{11.827} & \multirow{2}{*}{31.113} \\
 & & $b$ & $3.5937\times10^{-7}$ & $5.12\times10^{-8}$ & & & \\
\addlinespace
\multirow{2}{*}{Linear-log} & \multirow{2}{*}{$T = a + b \cdot \ln P$} & $a$ & $-431.4626$ & $88.0119$ & \multirow{2}{*}{0.8288} & \multirow{2}{*}{14.852} & \multirow{2}{*}{24.615} \\
 & & $b$ & $26.9574$   & $5.0018$  & & & \\
\addlinespace
\multirow{2}{*}{Exponential} & \multirow{2}{*}{$T = a \cdot e^{b \cdot P}$} & $a$ & $24.459$ & $6.830$ & \multirow{2}{*}{0.7844} & \multirow{2}{*}{16.668} & \multirow{2}{*}{588.914} \\
 & & $b$ & $5.0992\times10^{-9}$ & $1.07\times10^{-9}$ & & & \\
\addlinespace
\multirow{2}{*}{Log-log} & \multirow{2}{*}{$\ln T = a + b \cdot \ln P$} & $a$ & $-11.2501$ & $1.8077$  & \multirow{2}{*}{0.8969} & \multirow{2}{*}{11.524}  & \multirow{2}{*}{17.864} \\
 & & $b$ & $0.8269$   & $0.1027$  & & & \\
\bottomrule
\end{tabular}}
\caption{Comparison of four functional forms for predicting the onset of plasticity loss, $T$, as a function of effective parameter count, $P$. Coefficients are reported with their standard errors (STE). The in-sample root-mean-squared error (RMSE) was measured on the data points used to fit the model, whereas the out-of-sample RMSE was computed using leave-one-out cross-validation.}
\label{tab:model-comparison}
\end{table*}

\section{Correlates of Plasticity Loss in the Multilingual Learning Problems}
\label{app:correlates}

\begin{figure*}[t]
    \centering
    \includegraphics[width=\textwidth]{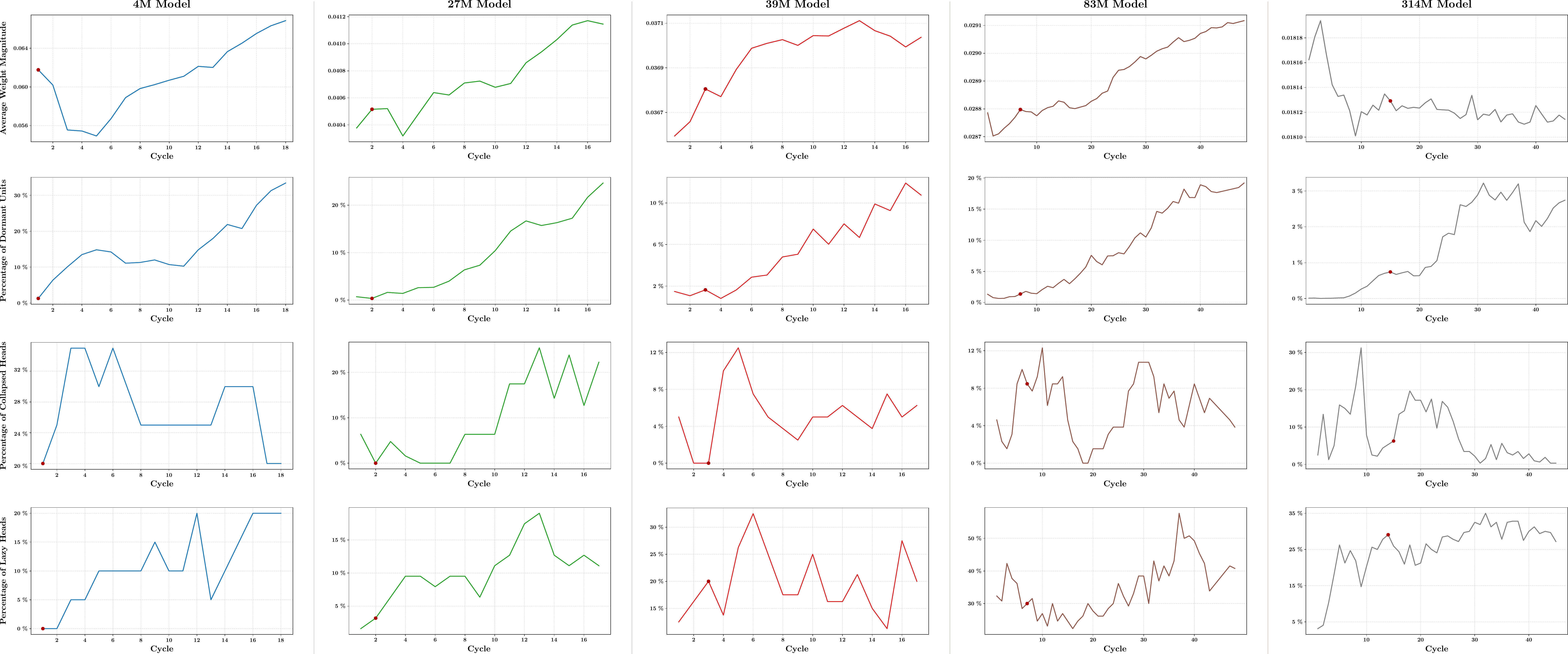}
    \caption{Average weight magnitude, percentage of dormant units, percentage of collapsed heads, and percentage of lazy heads for the 5M, 27M, 39M, and 83M models.}
    \label{fig:ext_corr}
\end{figure*}

In this appendix, we present extended results about the correlates of plasticity loss. In Figure \ref{fig:ext_corr}, we show the correlates of loss of plasticity in the Continual Multilingual Learning Problem for the models omitted in the main text: 5M, 27M, 39M, 83M, and 314M. The results are consistent with the discussion in the main text. While average weight magnitude tends to increase and the models often accumulate dormant units, lazy heads, and collapsed heads, these effects do not exactly match the deterioration in performance due to plasticity loss. 

\begin{figure*}[t]
    \centering
    \includegraphics[width=0.8\textwidth]{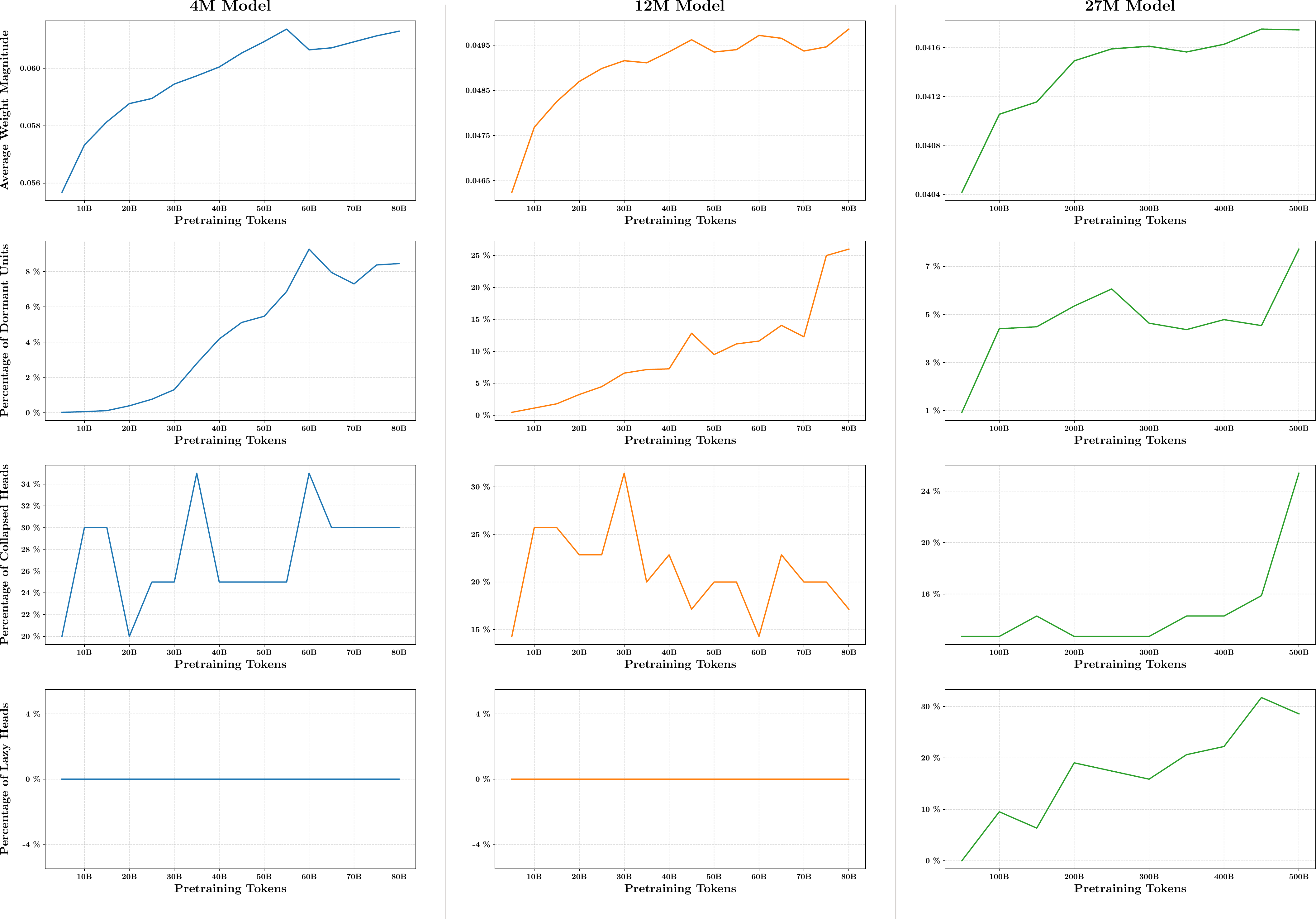}
    \caption{Correlates of loss of plasticity for the 5M, 12M, and 27M models trained on the Multilingual Stationary Learning Problem. The metrics followed a similar trend to the Multilingual Continual Learning Problem, but none of them exactly matched the trend in plasticity loss.}
    \label{fig:stat_corr}
\end{figure*}

Figure \ref{fig:stat_corr} shows the correlates of plasticity loss for the models trained in the Multilingual Stationary Learning Problem. Yet again, we saw an accumulation of dormant units and collapsed heads as the models lost plasticity. The percentage of lazy heads increased only in the 27M models, while the 5M and 12M models remained at zero percent collapsed heads. The lack of accumulation of collapsed heads is consistent with the results in the continual learning setting, where the 5M and 12M models had close to zero percent of collapsed heads during the first two cycles, i.e., the first 80 billion tokens of training. The average weight magnitude also increased over time, but did not seem to match the effect of loss of plasticity, as it continued to increase while performance was still improving in the probing tasks. 

Figures \ref{fig:ext_corr} and \ref{fig:stat_corr} emphasize the unreliability of the correlates of loss of plasticity. Nevertheless, they signal potential pathological changes happening inside the network that could be addressed through the approaches discussed in the main text. 

\end{document}